\documentclass[11pt]{article}

\usepackage[preprint]{acl}

\usepackage{newtxtext,newtxmath}
\usepackage{latexsym}
\usepackage[T1]{fontenc}
\usepackage[utf8]{inputenc}
\usepackage{microtype}
\usepackage{xspace}
\usepackage{graphicx}
\usepackage{booktabs}
\usepackage{array}
\usepackage{amsmath,mathtools}
\usepackage{enumitem}
\usepackage{xcolor}
\usepackage{multirow}
\usepackage{float}
\usepackage{makecell}
\usepackage{tabularx}

\newcommand{\ours}{\textsc{DualGuard}\xspace}
\title{DualGuard: Dual-Mode Quality Control for Logic-Preserving Data Augmentation}

\author{Shenghao Li \and Lin Zhao \\
  Guangzhou University, Guangzhou, China \\
  \texttt{lishenghao@e.gzhu.edu.cn}, \texttt{zhaolin@e.gzhu.edu.cn}}

\begin{document}
\maketitle

\begin{abstract}
Large language models provide a practical way to generate augmented data for logical reasoning at scale, but a larger generation volume does not guarantee semantic, label, or logical reliability. Existing work has improved generation quality through generation constraints, candidate validation, filtering, and feedback-based revision; however, once a quality judgment is available, deciding whether a candidate should be retained, filtered, or repaired remains an important control problem. We propose \ours, a dual-mode quality-control framework for logic-preserving data augmentation. The first mode uses the current instance and candidate batch for selective retention, filtering, attribution, and targeted feedback. The second mode accumulates cross-instance execution records of augmentation actions on top of per-sample diagnosis and attribution, compares new executions against each action's own historical behavior, and supports retrospective anomaly inspection, targeted rollback, and bounded repair. Both modes share semantic verification and additionally use symbolic verification when a reliable logical form is available. Across seven downstream tasks in the Two-Stage Transfer setting, \ours achieves the highest Accuracy on five tasks and outperforms the no-augmentation BERT baseline on all seven. Controlled ablations further show complementary roles for Memory, Z3, and history-aware anomaly control.
\end{abstract}

\section{Introduction}\label{sec:intro}
Recent advances in large language models (LLMs) have made it possible to synthesize training data at scale, offering a practical route to alleviate the scarcity of high-quality supervision for logical reasoning. Compared with ordinary text classification, logical reasoning examples must be linguistically natural while also preserving the inference relation and target label. LLM-based augmentation can expand training data through rewriting, attribute manipulation, task-specific instructions, and controlled generation \citep{yoo2021gpt3mix,zhou2022flipda,yu2023attrprompt,peng2024cotam,dai2025auggpt,li2024selfllmda}. Nevertheless, augmentation gains are not always stable and can vary with the generator, prompt design, downstream task, and data scale; in difficult or few-shot settings, augmentation may even provide only limited benefit or degrade performance \citep{zhou2022flipda,li2024selfllmda}.

This problem is especially acute in logical reasoning. Small changes to quantifiers, negation, implication direction, or premise content can alter the original reasoning relation or even flip the target label. Consequently, prior research has increasingly shifted from simply enlarging the generated set toward controlling and validating augmented data. Logic-driven methods introduce explicit logical or structured semantic constraints \citep{wang2022logicdriven,bao2024amrlda,li2026lfcda}; other work uses label-aware generation, task-specific instructions, or iterative enhancement. Logical DA further combines multi-agent generation, label verification, and reflection based on logical-validation feedback, allowing validation signals to affect later generation and revision \citep{zheng2025logicalda}.

Existing work therefore already provides generation constraints, candidate validation, filtering, and feedback-based revision. Our focus is one step later: after a quality signal has been obtained, how should the system decide whether a particular candidate should be retained, filtered, or repaired? The current instance and candidate batch summarize the state of the present generation round, whereas cross-instance histories of augmentation actions provide a behavioral reference for the current execution and for reviewing recent historical anomalies.

We address this problem with \ours, an agentic dual-mode quality-control framework for logic-preserving augmentation. The first mode uses validation outcomes, candidate scores, risk information, and within-batch stability from the current instance to selectively retain reliable candidates, filter problematic ones, and perform attribution and targeted feedback when a diagnosable issue is present. The second mode first performs per-sample diagnosis and attribution, then accumulates action-level execution rewards and diagnostic records across samples. Each action's own historical behavior becomes a dynamic reference for identifying abnormal current or recent executions, which can trigger targeted rollback and bounded repair. The two modes share semantic verification and conditionally combine it with symbolic logic verification when a reliable logical representation can be obtained. The attribution and rollback process follows the supervision-and-coordination idea of multi-agent systems by routing corrective information to the processing node most likely responsible for the observed failure.

We evaluate \ours on multiple logical reasoning and natural-language inference tasks. In the Two-Stage Transfer setting, \ours obtains the highest Accuracy on five of seven downstream tasks and exceeds the no-augmentation BERT baseline on all seven. Controlled ablations, direct generation-quality evaluation, and process-level analyses further examine the behavior and contribution of the major components.

Our main contributions are:
\begin{enumerate}
    \item \textbf{A dual-mode quality-control framework for post-generation decisions.} \ours refines post-verification processing into retention, filtering, and repair, and pre-configures either current-batch control or history-aware control for a task so that validation results can directly guide candidate handling and subsequent generation.
    \item \textbf{Attribution-based targeted rollback and bounded repair.} For diagnosable failures, \ours combines candidate state, validation results, quality scores, and diagnostic signals to localize likely failure sources and route corrective feedback only to the relevant processing node, rather than restarting the entire generation process.
    \item \textbf{Cross-instance experience memory.} Memory continuously stores augmentation actions together with their cross-sample execution outcomes, rewards, and diagnostics so that reusable experience can support later samples. History-aware anomaly control then uses these records to build action-specific dynamic references for current decisions and retrospective anomaly inspection.
\end{enumerate}

\section{Related Work}\label{sec:related}
\noindent\textbf{Language-Model-Based Data Augmentation.}
Text augmentation has evolved from lightweight token-level perturbations to controllable generation with pretrained and instruction-following language models. EDA uses synonym replacement, random insertion, swapping, and deletion \citep{wei2019eda}, while Contextual Augmentation replaces words using contextual language models \citep{kobayashi2018contextual}. GPT3Mix uses large language models for synthetic text augmentation \citep{yoo2021gpt3mix}. FlipDA improves few-shot augmentation through label-flipped generation \citep{zhou2022flipda}; AttrPrompt exposes explicit attributes for controllable generation \citep{yu2023attrprompt}; and CoTAM uses chain-of-thought-guided attribute manipulation \citep{peng2024cotam}. AugGPT and Self-LLMDA further demonstrate the potential of instruction-driven LLMs for task-specific augmentation \citep{dai2025auggpt,li2024selfllmda}.

\noindent\textbf{Logic-Aware Augmentation and Verification.}
As logical consistency becomes a stronger requirement, several methods explicitly incorporate logical structure or structured semantic representations into augmentation. Logic-Driven Context Extension and Data Augmentation constructs logically related variants for reasoning supervision \citep{wang2022logicdriven}; AMR-based augmentation uses structured semantic representations to guide logical transformations \citep{bao2024amrlda}; and logic-formulated control introduces explicit logical constraints into natural-language reasoning augmentation \citep{li2026lfcda}. Logical DA combines multi-agent generation, label verification, and reflection based on logical-validation feedback \citep{zheng2025logicalda}.

Post-generation validation and feedback are also central to reliable generation. Self-Refine iteratively improves outputs using self-generated feedback \citep{madaan2023selfrefine}, while CRITIC uses external tools for interactive critique and correction \citep{gou2024critic}. In parallel, neuro-symbolic reasoning systems such as Logic-LM, SatLM, and Faithful Chain-of-Thought translate natural-language reasoning into more explicit or executable forms and rely on external solvers or constrained reasoning to improve faithfulness \citep{pan2023logiclm,ye2023satlm,lyu2023faithfulcot}. These studies show that validation signals can serve not only as rejection criteria but also as inputs to later feedback and revision.

\noindent\textbf{Feedback, Memory, and Historical Experience.}
Feedback and memory have increasingly been used to guide repeated generation and decision making. LLM2LLM uses errors exposed by a student model to drive targeted synthetic-data generation by a teacher model \citep{lee2024llm2llm}. R-GDA employs a multi-agent feedback loop to refine augmentation guidance and calibrate generated data \citep{kang2026rgda}. Beyond data augmentation, ReAct interleaves reasoning and action \citep{yao2023react}; Reflexion converts feedback into verbal reflections that are stored for later trials \citep{shinn2023reflexion}; ExpeL extracts reusable experience from previous tasks \citep{zhao2024expel}; and MemoryBank studies persistent long-term memory for language models \citep{zhong2024memorybank}.

Building on these lines of work, \ours focuses specifically on the control process after candidate quality has been assessed and distinguishes two sources of information: the state of the current instance and candidate batch, and the cross-instance historical behavior of augmentation actions. The former supports selective retention, filtering, and handling of diagnosable problems in the current batch. The latter is accumulated on top of per-sample diagnosis and attribution and serves as an action-specific dynamic reference for detecting abnormal current or recent executions and for supporting targeted rollback and bounded repair.

\section{Method}\label{sec:method}
\subsection{Overall Framework}\label{sec:overall}
\begin{figure*}[t]
\centering
\includegraphics[width=0.94\textwidth]{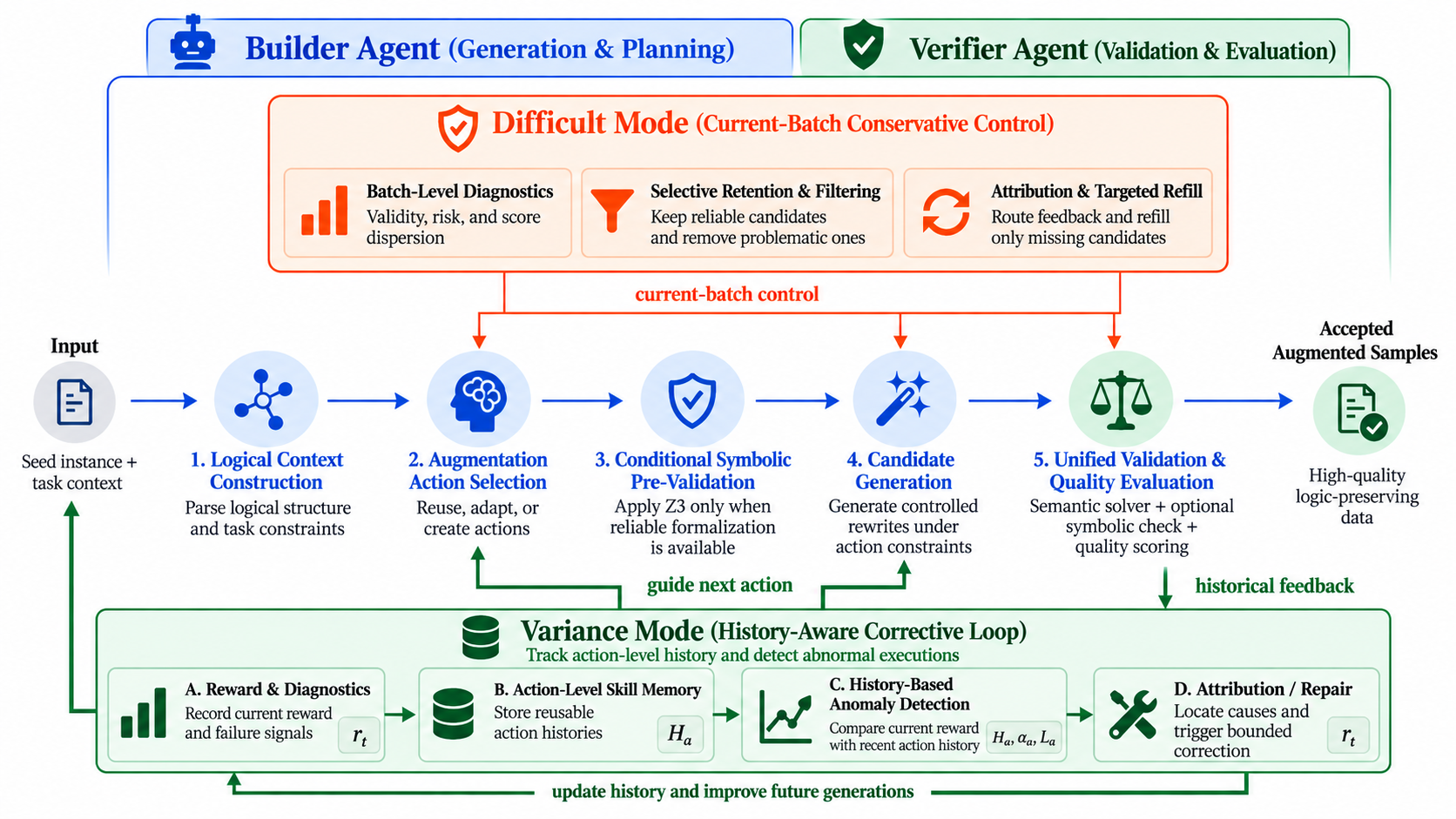}
\caption{Overall framework of \ours. The shared generation and validation workflow feeds one task-preconfigured quality-control mode: current-batch control in Difficult Mode or history-aware control in Variance Mode.}
\label{fig:framework}
\end{figure*}

Figure~\ref{fig:framework} presents the overall workflow. \ours pre-configures the appropriate quality-control mode for a dataset or task and performs logic-preserving augmentation within a shared generation-and-validation framework. Given a reasoning instance, the system first analyzes its logical structure, entity relations, and task constraints, and determines whether the instance can be reliably formalized. A controller then selects, reuses, or constructs an augmentation action using the current instance and existing action information. Each action is defined by an action name, an explicit rewriting rule, and a natural-language description, and may additionally store applicability metadata such as logic type, label, and historical performance. The Builder applies the selected rule and description to produce candidate samples.

Generated candidates then enter a shared validation and quality-assessment module. The module conditionally enables symbolic verification according to formalizability and combines it with a semantic Solver that checks logical, semantic, and label consistency. It outputs candidate validity, quality scores, and structured diagnostics. Candidates satisfying basic validity continue to the mode-specific control process, while invalid candidates are filtered. Section~\ref{sec:validation} formalizes this shared validation mechanism.

In \textbf{Difficult Mode}, the system primarily uses validation outcomes, quality scores, risk information, and within-batch stability from the current instance and candidate batch. Reliable candidates are retained, diagnosable failures are filtered and attributed, corrective information is routed to the relevant node, and only missing candidates are regenerated. Section~\ref{sec:difficult} gives the detailed scoring and batch-control procedure.

In \textbf{Variance Mode}, the system records action-level execution outcomes on top of per-sample diagnosis and attribution, including reward, diagnostics, and problem type. As cross-instance executions of the same action accumulate, the action's historical behavior defines a dynamic reference for identifying abnormal current or recent executions. When necessary, the system uses the already-available diagnosis and attribution to perform historical feedback, targeted rollback, and bounded repair. Section~\ref{sec:variance} details the history-aware statistics and control rule.

The two modes are pre-configured alternatives for different datasets or tasks rather than two sequential stages through which every candidate must pass. Difficult Mode is driven mainly by the current instance and current batch, whereas Variance Mode additionally exploits cross-instance historical behavior of augmentation actions. Both modes share the generation and validation components described above.

\subsection{Shared Candidate Validation}\label{sec:validation}
Both modes share the same candidate-validation mechanism. For an input instance $x$, the system first determines whether a reliable logical representation is available:
\begin{equation}
\rho(x)=
\begin{cases}
1, & \text{if $x$ can be reliably formalized},\\
0, & \text{otherwise}.
\end{cases}
\end{equation}
When $\rho(x)=1$, the available representation is considered sufficient for symbolic relation checking with the Z3 SMT solver \citep{demoura2008z3}. When $\rho(x)=0$, the system does not force a symbolic conversion and skips Z3 verification.

For a generated candidate $c$, the semantic Solver checks task-dependent semantic and label constraints, including label consistency, required semantic preservation, entity correspondence, polarity, logical direction, and expression completeness. Let
\begin{equation}
V_{\mathrm{sem}}(c)\in\{0,1\}
\end{equation}
be the semantic validation result. A value of 1 indicates that the candidate passes semantic validation.

When $\rho(x)=1$, the candidate is first subjected to Z3-based symbolic verification to check the logical relation that must be preserved. Only candidates passing this symbolic gate proceed to the semantic Solver; symbolic failures are immediately marked invalid and discarded rather than routed to later attribution or repair. When $\rho(x)=0$, the system skips Z3 and directly applies semantic validation. Let
\begin{equation}
V_{\mathrm{Z3}}(c)\in\{0,1\}
\end{equation}
be the symbolic validation result. The basic validity of a candidate can then be written as
\begin{equation}
V(c)=V_{\mathrm{sem}}(c)\left[(1-\rho(x))+\rho(x)V_{\mathrm{Z3}}(c)\right].
\end{equation}
Thus, if $\rho(x)=0$, $V(c)=V_{\mathrm{sem}}(c)$; if $\rho(x)=1$, $V(c)=V_{\mathrm{sem}}(c)V_{\mathrm{Z3}}(c)$. Z3 is therefore a conditionally enabled symbolic safeguard, while the semantic Solver remains part of the basic validity decision for every candidate that reaches it. The validation stage also outputs candidate scores and structured diagnostics for subsequent mode-specific control.

\subsection{Difficult Mode: Current-Batch Quality Control}\label{sec:difficult}
\begin{figure*}[t]
\centering
\includegraphics[width=0.96\textwidth]{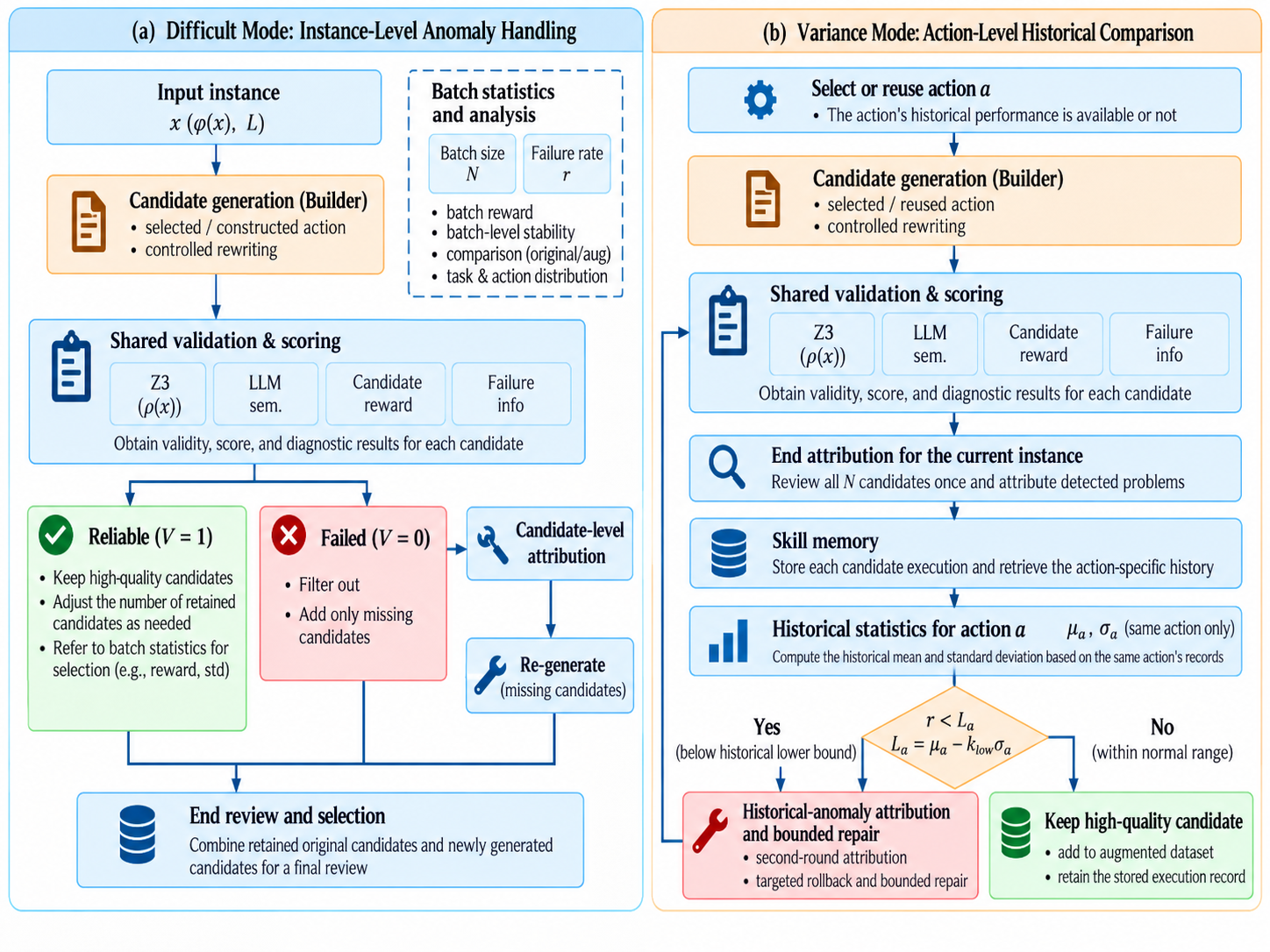}
\caption{Detailed dual-mode quality-control flow. (a) Difficult Mode uses the current instance and current candidate batch for selective retention, filtering, attribution, replenishment, and final review. (b) Variance Mode performs per-sample diagnosis first and then uses action-level historical behavior to detect low historical anomalies and support targeted repair.}
\label{fig:dualmode}
\end{figure*}

Figure~\ref{fig:dualmode}(a) summarizes Difficult Mode. It controls the current reasoning instance and its current candidate batch without relying on cross-instance historical memory. Let the candidate set generated at round $t$ be
\begin{equation}
C_t=\{c_1,c_2,\ldots,c_n\}.
\end{equation}
Candidates first pass through the shared validation in Section~\ref{sec:validation}. A candidate that requires symbolic verification but fails Z3 is immediately discarded and does not enter subsequent attribution or repair. Remaining candidates are assessed using individual quality, risk information, and current-batch state.

For candidate $c_i$, let $f_i$ denote Solver fluency and $d_i$ its confusion score. The individual candidate score is
\begin{equation}
s_i=f_i-P_{\mathrm{fluency}}-P_{\mathrm{risk}}-P_{\mathrm{lowvalue}}-P_{\mathrm{confusion}},
\end{equation}
with fluency penalty
\begin{equation}
P_{\mathrm{fluency}}=\max(0,\tau_f-f_i).
\end{equation}
Let $H_i,L_i\in\{0,1\}$ indicate high-risk behavior and low-value rewriting, respectively. Their penalties are
\begin{equation}
P_{\mathrm{risk}}=\lambda_{\mathrm{risk}}H_i,\qquad
P_{\mathrm{lowvalue}}=\lambda_{\mathrm{low}}L_i.
\end{equation}
Difficult Mode does not directly reward a higher confusion score. It only penalizes confusion beyond an allowed range:
\begin{equation}
P_{\mathrm{confusion}}=\lambda_{\mathrm{conf}}\max(0,d_i-\tau_{\mathrm{conf}}).
\end{equation}
The score therefore summarizes language quality, risk, rewriting value, and difficulty control. These thresholds and penalty weights remain fixed across the reported Difficult Mode experiments.

The system also analyzes the state of the candidates that actually remain for current-round batch statistics. Let $N_t$ be this count. Their mean score and population standard deviation are
\begin{equation}
\bar{s}_t=\frac{1}{N_t}\sum_{i=1}^{N_t}s_i,
\end{equation}
\begin{equation}
\sigma_{s,t}=\sqrt{\frac{1}{N_t}\sum_{i=1}^{N_t}(s_i-\bar{s}_t)^2}.
\end{equation}
The system additionally considers the valid-candidate ratio, fluency variation, and risk information. A large $\sigma_{s,t}$ only indicates that candidate quality varies substantially within the batch; it is not by itself evidence of a logical error, so batch statistics are used only as auxiliary quality signals.

After individual scoring and batch-state analysis, the system combines shared validity, candidate score, risk state, and batch information into a current-mode quality-control decision
\begin{equation}
G_D(c_i,C_t)\in\{0,1\}.
\end{equation}
The retained set is
\begin{equation}
A_t=\{c_i\in C_t:V(c_i)=1,\;G_D(c_i,C_t)=1\},
\end{equation}
and rejected candidates are
\begin{equation}
F_t=C_t\setminus A_t.
\end{equation}
Once a candidate enters $A_t$, it is retained even if another member of the same batch fails.

For failed candidates with explicit semantic, generation, or risk diagnostics, the system performs per-candidate attribution. The attribution module combines Solver judgments, candidate scores, risk information, and failure reasons to determine whether the issue most likely originates from logic-context construction, action selection, Builder generation, or another repairable node, and produces targeted feedback accordingly.

Importantly, attribution is not triggered solely by whether the within-batch score standard deviation is abnormal. Any candidate with an explicit diagnosable problem can be attributed even when the overall batch appears stable. After filtering and attribution, reliable candidates remain in place and the system generates only the missing number of replacements. The retained and newly generated candidates then undergo a final End-stage batch review. Previously filtered candidates that have already been attributed are not processed again; only newly detected problems from the End review receive additional attribution and correction.

\subsection{Variance Mode: History-Aware Quality Control}\label{sec:variance}
Figure~\ref{fig:dualmode}(b) summarizes Variance Mode. The mode analyzes the current execution and recent history using cross-instance execution records of augmentation actions applicable to the current reasoning type. For each instance, the system identifies the relevant logic type and retrieves its action-level Skill Memory. The action-selection agent may reuse an existing action, create a new one, or explore both according to configuration. After generation and shared validation, the system first performs per-sample diagnosis and records validity, risk signals, and problem type.

The system then calculates a reward for the current action execution. For candidate $c$,
\begin{equation}
r(c)=\alpha C(c)-P_{\mathrm{novelty}}(c)-P_{\mathrm{fluency}}(c),
\end{equation}
where $C(c)$ is the Solver confusion score and $\alpha$ controls its contribution. Insufficient fluency incurs
\begin{equation}
P_{\mathrm{fluency}}(c)=\max(0,\tau_f-F(c)),
\end{equation}
where $F(c)$ is candidate fluency. Novelty penalty is defined as
\begin{equation}
\begin{split}
P_{\mathrm{novelty}}(c)=&\;\lambda_n\left(w_sS_{\mathrm{style}}+w_lS_{\mathrm{logic}}+w_cS_{\mathrm{combo}}\right)\\
&+\lambda_a\log(1+N_a),
\end{split}
\end{equation}
where $S_{\mathrm{style}}$ is style similarity, $S_{\mathrm{logic}}$ is logical-label-structure similarity, $S_{\mathrm{combo}}$ indicates repetition of the complete style--logic pattern, and $N_a$ is the historical reuse count of action $a$. Style similarity is computed by embedding style-label descriptions with a local text encoder; logical similarity is obtained from overlap among logic labels; and the combination term records whether the full style--logic pattern has already occurred. For a reused action, records previously produced by that same action are excluded from the similarity reference, while its reuse frequency is penalized separately through $N_a$.

These configurable dimensions, weights, and thresholds express task-specific preferences among candidates that have already satisfied the basic logical and semantic validity constraints. The experiments use a fixed set of dimensions and parameters. Using the diagnostic signals together with the computed reward, the system then performs per-sample attribution and records the likely source before updating memory.

After the current execution, the system stores the action reward, diagnostics, problem type, and related execution information in the corresponding Skill Memory. To bound memory size and keep the statistics recent, each action within a logic type maintains a sliding window of length $W$:
\begin{equation}
\mathcal{R}_a=\{r_1,r_2,\ldots,r_n\},\qquad n\le W.
\end{equation}
The historical mean and standard deviation are
\begin{equation}
\mu_a=\frac{1}{n}\sum_{i=1}^{n}r_i,
\end{equation}
\begin{equation}
\sigma_a=\sqrt{\frac{1}{n}\sum_{i=1}^{n}(r_i-\mu_a)^2}.
\end{equation}
Both statistics are action-specific and summarize the recent behavior of the same action on compatible samples; they are not used to compare absolute reward values across different actions.

The system defines an action-specific dynamic lower bound
\begin{equation}
L_a=\mu_a-k_{\mathrm{low}}\sigma_a.
\end{equation}
A current execution with reward $r_t$ is marked as a low historical anomaly when
\begin{equation}
r_t<L_a.
\end{equation}
Such an anomaly does not by itself prove a logical error. It indicates that the current execution is substantially worse than the action's own recent behavior, making the deviation a routing signal rather than a universal quality threshold.

Historical anomaly checking is enabled only after enough evidence has accumulated:
\begin{equation}
I_a=\mathbb{I}[n\ge n_{\min}]\cdot\mathbb{I}[\sigma_a\ge\sigma_{\min}].
\end{equation}
For a new action with insufficient history, the system only accumulates rewards and diagnostics and does not yet make a history-based anomaly judgment.

When a low historical anomaly is detected, the system combines the current reward, action-history statistics, and the diagnosis and attribution that were already obtained for the sample to identify the node requiring additional processing. It then performs targeted rollback and repair under a bounded repair budget. Repaired candidates re-enter shared validation and are accepted only if they again satisfy the logical, semantic, and current quality requirements. The sliding window also supports retrospective inspection of recent executions: if a stored record appears abnormal under the current dynamic reference, its associated diagnostics can be revisited to guide subsequent processing. Variance Mode therefore uses an action's own recent history both as a reference for the current execution and as evidence for identifying recent instability.

\section{Experimental Setup}\label{sec:setup}
\noindent\textbf{Datasets and Task Protocols.}
We use two evaluation protocols: \textbf{Two-Stage Transfer} and \textbf{Direct Augmentation}. The former first trains a logic-oriented intermediate model on augmented data and then transfers it to downstream tasks; the latter augments the target-task training data directly and evaluates the resulting downstream model.

For Two-Stage Transfer, we construct a general logic seed set of 150 premise--conclusion pairs, consisting of 75 valid and 75 invalid inferences. The set covers five basic logical structures---conditional reasoning, contraposition, conjunction, syllogistic reasoning, and disjunction---with 15 valid and 15 invalid examples per type. Each example contains premises, a conclusion, and a logical-validity label. Each augmentation method generates data from the same 150 seeds and trains its own Stage-1 intermediate model. These models are then transferred to seven downstream tasks: FOILO, ProofWriter, LogicQA, ReClor, RTE, CB, and LogDed. ProofWriter is a rule-based natural-language reasoning benchmark \citep{tafjord2021proofwriter}; the local FOILO resource follows the first-order-logic setting represented by FOLIO \citep{han2024folio}; ReClor and LogiQA/LogicQA target logical reading comprehension \citep{yu2020reclor,liu2020logiqa}; RTE and CB are SuperGLUE tasks \citep{wang2019superglue}; and LogDed follows logical-deduction evaluation in the BIG-bench family \citep{srivastava2023bigbench}.

Direct Augmentation omits Stage-1 intermediate-model training. It directly augments original target-task training examples and trains the downstream classifier on original plus augmented data. This protocol includes CB, RTE, FOILO, and ProofWriter, with 48, 100, 150, and 150 original examples used as augmentation sources, respectively.

We follow the official splits available for each task. For FOILO, ReClor, RTE, and CB, the official validation set is reserved for final evaluation and 10\% of the original training split is held out for model selection. ProofWriter uses its official test split for final evaluation and likewise holds out 10\% of training data for validation. LogicQA and LogDed retain their official train/validation/test organization. All methods use identical splits. In Direct Augmentation, examples are grouped by \texttt{source\_id} so that an original example and all of its derived augmentations remain within the same partition, preventing source-level leakage.

\noindent\textbf{Baselines.}
We use BERT-base without data augmentation as the reference model \citep{devlin2019bert} and compare against four representative augmentation baselines: AugGPT \citep{dai2025auggpt}, AttrPrompt \citep{yu2023attrprompt}, CoTAM \citep{peng2024cotam}, and FlipDA \citep{zhou2022flipda}. We implement each method using prompts and augmentation strategies that follow the corresponding method's core design.

For fairness, all augmentation methods use the same original examples, generator, decoding temperature, and generation budget; only the prompt templates and augmentation operations required by each method differ. Generated candidates are quality-checked before downstream training. Consequently, retained augmentation counts may differ slightly across methods, but the overall data scales remain comparable. Candidates failing label consistency, semantic quality, duplication, or logical validity checks are filtered rather than replaced by duplicate examples or by relaxing the quality standard.

\noindent\textbf{Generation, Training, and Evaluation.}
All augmentation methods use \texttt{agnes-2.0-flash} for candidate generation at a temperature of 0.6. In \ours, logic-structure extraction, candidate generation, semantic evaluation, and feedback attribution share the same model configuration and temperature. All methods operate under the same generation budget. \ours additionally constrains candidate generation and repair by a fixed budget; candidates that still fail quality requirements are discarded rather than force-filled.

All downstream classifiers are initialized from the same BERT-base-uncased checkpoint \citep{devlin2019bert}. Training uses AdamW, linear learning-rate warmup and decay, gradient clipping, validation-based model selection, and early stopping. In Two-Stage Transfer, each augmentation method first trains its own Stage-1 intermediate model and selects the best Stage-1 configuration within the predefined range. For each downstream task in Stage 2, we randomly select one configuration of non-learning-rate hyperparameters and use it for every method. We then vary only the learning rate across the same three predefined values and report the arithmetic mean of the three Accuracy results. All methods and learning-rate configurations use the same fixed random seed within each task, while the other Stage-2 training settings are also held constant across methods.

In Direct Augmentation, each method directly trains on original plus augmented target-task data under a fixed three-fold cross-validation protocol with a shared random seed. All methods share the same folds and training setup, and augmented examples are added only to the corresponding training fold. External test data are never used for hyperparameter tuning, model selection, or early stopping. Accuracy is the primary metric: Two-Stage Transfer reports the mean across the three predefined learning-rate configurations, while Direct Augmentation reports the mean across the three fixed validation folds.

\noindent\textbf{Implementation and Hardware.}
The data-generation and multi-node workflow is implemented in Python and LangGraph. LLM calls use a remote OpenAI-compatible API for action selection, candidate generation, semantic evaluation, and feedback attribution. Z3 Solver performs symbolic verification whenever a candidate can be reliably formalized. In Variance Mode, style similarity is computed by embedding style-label descriptions with local \texttt{all-MiniLM-L6-v2}; logic similarity and combination repetition are computed from labels and rule statistics. Downstream BERT models are trained and evaluated with PyTorch and Transformers. Experiments use two NVIDIA GeForce RTX 3090 GPUs with 24~GB memory each; GPUs are primarily used for downstream training and evaluation, while LLM inference is served remotely.

\section{Results and Analysis}\label{sec:results}
We analyze \ours from four complementary perspectives: downstream performance, component contribution, generated-data quality, and process-level control behavior.

\noindent\textbf{Two-Stage Transfer Results.}
Table~\ref{tab:twostage} reports Accuracy on the seven downstream tasks. \ours obtains the highest Accuracy on FOILO, ProofWriter, LogicQA, ReClor, and LogDed and exceeds the no-augmentation BERT baseline on all seven tasks.

\begin{table*}[t]
\centering
\small
\setlength{\tabcolsep}{4.2pt}
\renewcommand{\arraystretch}{1.08}
\caption{Two-Stage Transfer Accuracy (\%), averaged over three predefined learning rates using the same fixed random seed across methods. Best result per task is bold.}
\label{tab:twostage}
\begin{tabular}{lrrrrrrr}
\toprule
Method & FOILO & ProofWriter & LogicQA & ReClor & RTE & CB & LogDed \\
\midrule
BERT       & 51.89 & 52.87 & 29.70 & 46.20 & 65.22 & 85.12 & 17.33 \\
AugGPT     & 50.57 & 59.63 & 30.41 & 47.13 & 69.80 & 83.93 & 24.00 \\
AttrPrompt & 56.27 & 56.94 & 30.62 & 48.13 & 70.04 & 85.71 & 21.33 \\
CoTAM      & 56.65 & 54.81 & 29.95 & 47.73 & \textbf{71.24} & 87.50 & 20.89 \\
FlipDA     & 53.04 & 56.57 & 29.54 & 49.40 & 70.16 & \textbf{90.48} & 24.89 \\
\ours     & \textbf{57.47} & \textbf{61.02} & \textbf{31.49} & \textbf{49.67} & 69.92 & 86.31 & \textbf{25.67} \\
\bottomrule
\end{tabular}
\end{table*}

\begin{figure*}[t]
\centering
\includegraphics[width=0.72\textwidth]{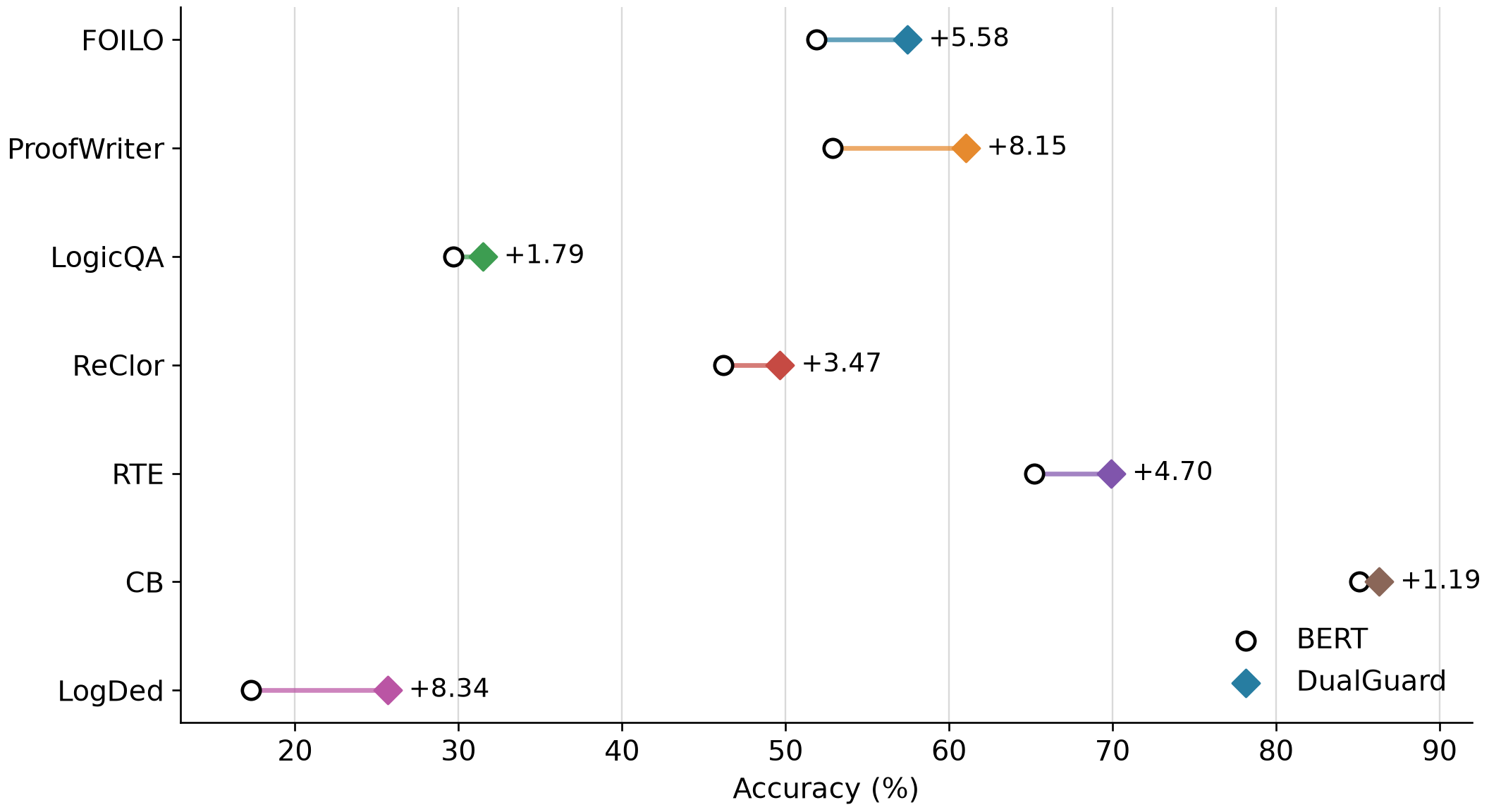}
\caption{BERT versus \ours in Two-Stage Transfer. Labels show the absolute Accuracy gain of \ours over BERT for each task.}
\label{fig:twostage}
\end{figure*}

Relative to BERT, the gains on LogDed, ProofWriter, and FOILO are 8.34, 8.15, and 5.58 percentage points, respectively, and the mean gain across all seven tasks is 4.75 points. CoTAM and FlipDA remain strongest on RTE and CB, respectively, showing that augmentation benefits are task-dependent rather than universally dominated by a single strategy. Figure~\ref{fig:twostage} visualizes the absolute BERT-to-\ours changes.

\noindent\textbf{Direct Augmentation Results.}
Direct Augmentation evaluates augmentation directly on the target-task data distribution. Table~\ref{tab:direct} reports fixed three-fold mean Accuracy for the four tasks.

\begin{table*}[t]
\centering
\small
\setlength{\tabcolsep}{8pt}
\renewcommand{\arraystretch}{1.08}
\caption{Direct Augmentation Accuracy (\%), averaged over three fixed validation folds. Best result per task is bold; FOILO has a three-way tie.}
\label{tab:direct}
\begin{tabular}{lrrrr}
\toprule
Method & FOILO & ProofWriter & RTE & CB \\
\midrule
BERT       & 42.67 & 45.33 & 50.00 & 70.83 \\
AugGPT     & 43.33 & 48.67 & 53.00 & 75.00 \\
AttrPrompt & \textbf{44.67} & 53.33 & 54.99 & 72.92 \\
CoTAM      & 43.33 & 38.00 & \textbf{66.01} & 75.00 \\
FlipDA     & \textbf{44.67} & 41.33 & 64.05 & 77.08 \\
\ours     & \textbf{44.67} & \textbf{54.67} & 57.04 & \textbf{79.17} \\
\bottomrule
\end{tabular}
\end{table*}

On ProofWriter and CB, \ours ranks first at 54.67\% and 79.17\%. It ties with AttrPrompt and FlipDA on FOILO (44.67\%). On RTE, it reaches 57.04\%, outperforming BERT, AugGPT, and AttrPrompt but remaining below CoTAM and FlipDA. Thus, the relative benefit of augmentation varies by target task rather than following the Two-Stage ranking.

\noindent\textbf{Controlled Ablation.}
To analyze the role of individual components, we compare the Full system against variants without Memory, Z3, or Variance under a separate controlled configuration. The absolute values in Table~\ref{tab:ablation} are interpreted only within this ablation protocol and are not substituted for the main Two-Stage results.

\begin{table*}[t]
\centering
\small
\setlength{\tabcolsep}{4.8pt}
\renewcommand{\arraystretch}{1.08}
\caption{Controlled ablation Accuracy (\%).}
\label{tab:ablation}
\begin{tabular}{lrrrrrrr}
\toprule
Variant & FOILO & ProofWriter & LogicQA & ReClor & RTE & CB & LogDed \\
\midrule
Full          & \textbf{59.28} & \textbf{61.02} & \textbf{31.49} & \textbf{50.20} & \textbf{71.08} & \textbf{85.12} & \textbf{25.78} \\
w/o Memory    & 57.97 & 57.78 & 29.24 & 45.73 & 70.95 & 85.12 & 22.67 \\
w/o Z3        & 57.31 & 58.98 & 28.42 & 46.13 & 69.61 & 85.12 & 21.78 \\
w/o Variance  & 59.11 & 56.85 & 29.60 & 47.07 & 66.93 & 80.36 & 24.89 \\
\bottomrule
\end{tabular}
\end{table*}

Full outperforms w/o Variance on all seven tasks, with drops ranging from 0.17 to 4.76 percentage points and averaging 2.74 points. The largest differences appear on CB, ProofWriter, and RTE. Removing Memory or Z3 also reduces Accuracy on most tasks, with mean drops of 2.07 and 2.37 points, respectively. These patterns indicate complementary roles for cross-instance experience, conditional symbolic verification, and action-level historical anomaly control. Because the current study does not include a standalone w/o Difficult Mode variant, we do not attribute an independently identified causal Accuracy gain to Difficult Mode itself.

\begin{figure*}[t]
\centering
\includegraphics[width=0.72\textwidth]{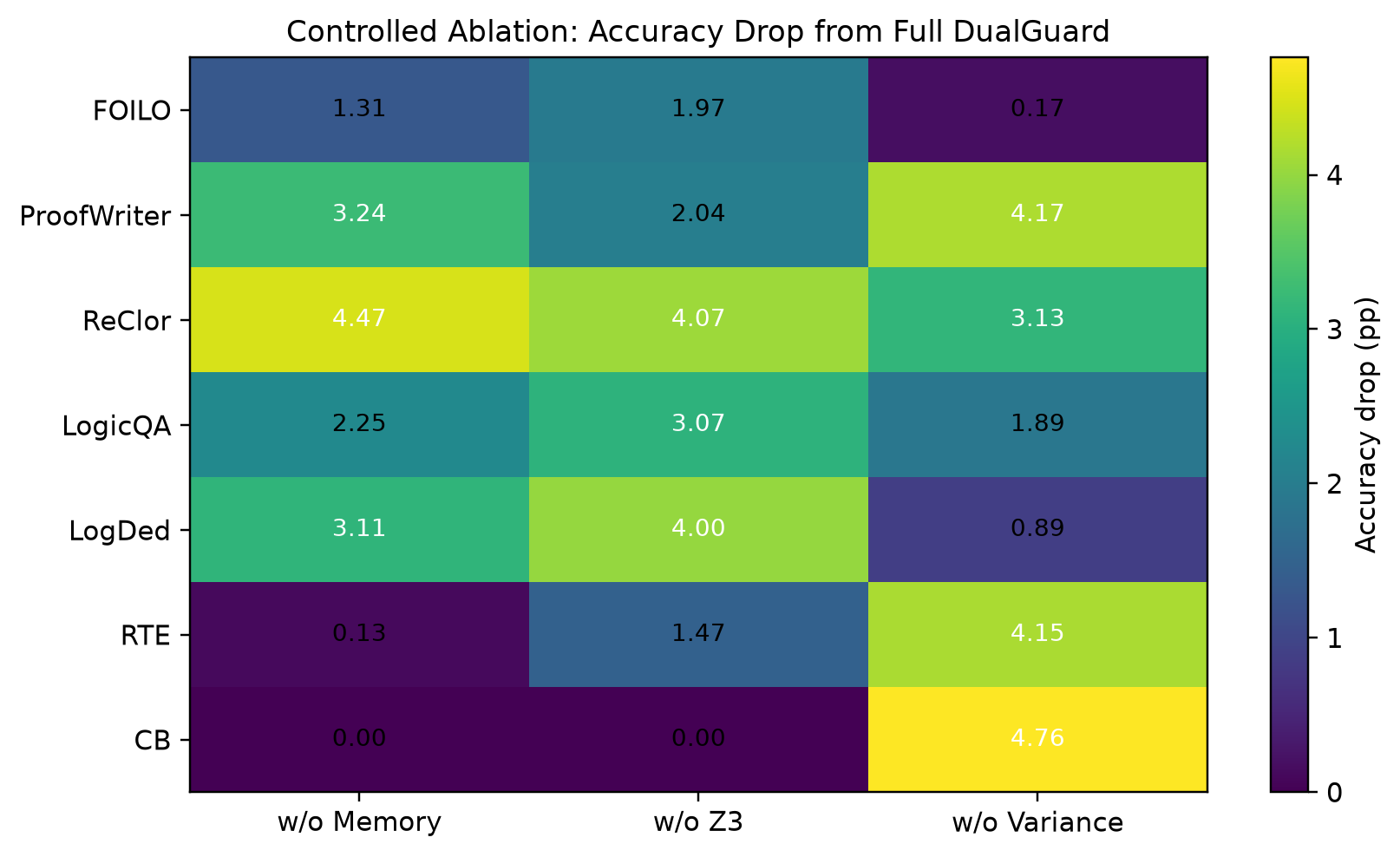}
\caption{Accuracy drop relative to the Full controlled ablation system after removing Memory, Z3, or Variance.}
\label{fig:ablation}
\end{figure*}

\noindent\textbf{Generated-Data Quality.}
We additionally conduct a small direct evaluation of the generated samples. For each of CB, FOILO, ProofWriter, RTE, and Simple Logic, we randomly sample 25 augmented examples from those validated and ultimately retained by the system, for a total of 125 samples. GPT-5.5 evaluates sample validity, logic-and-label correctness, transformation validity, language quality, and meaningful difference in this separate quality audit. This analysis describes the quality of \ours samples only and is not used to rank augmentation methods.

\begin{table*}[t]
\centering
\small
\setlength{\tabcolsep}{7pt}
\renewcommand{\arraystretch}{1.08}
\caption{Quality of selected \ours augmentations (25 per dataset). The first three metrics are pass counts; the last two are mean ratings on a five-point scale.}
\label{tab:quality}
\begin{tabular}{lccccc}
\toprule
Dataset & Sample validity & Logic/label & Transformation & Language & Difference \\
\midrule
CB           & 25/25 & 25/25 & 25/25 & 4.96 & 3.45 \\
FOILO        & 25/25 & 25/25 & 25/25 & 4.92 & 4.05 \\
ProofWriter  & 25/25 & 25/25 & 25/25 & 4.95 & 3.75 \\
RTE          & 25/25 & 25/25 & 25/25 & 4.96 & 3.35 \\
Simple Logic & 25/25 & 25/25 & 25/25 & 4.88 & 4.10 \\
\bottomrule
\end{tabular}
\end{table*}

All 125 selected examples pass the first three checks. Mean language-quality scores range from 4.88 to 4.96 out of 5, while meaningful-difference scores range from 3.35 to 4.10. This small audit describes the selected outputs, not the acceptance rate of all generated candidates or a cross-method quality ranking.

\noindent\textbf{Dual-Mode Process Analysis.}
Downstream Accuracy and ablations primarily measure final outcomes. We therefore also examine whether the two control modes behave as intended at the process level. Difficult Mode focuses on selective retention and replenishment in the current batch, whereas Variance Mode focuses on historical anomaly detection and subsequent repair.

\begin{table*}[t]
\centering
\small
\setlength{\tabcolsep}{5pt}
\caption{Difficult Mode on eight naturally triggered FOILO instances (from 24 traceable runs). Rows describe different candidate sets, not paired before--after repairs.}
\label{tab:difficultprocess}
\begin{tabular}{@{}lrrrr@{}}
\toprule
Candidate set & $n$ & Mean reward & Logic pass & High risk \\
\midrule
Initial candidates & 30 & 6.86 & 86.67\% & 13.33\% \\
Low-reward outliers & 8 & 1.59 & 50.00\% & 50.00\% \\
Attribution-guided refills & 35 & 8.35 & 91.43\% & 8.57\% \\
Final accepted & 32 & 9.09 & 100.00\% & 0.00\% \\
\bottomrule
\end{tabular}
\end{table*}

Among 24 FOILO runs with one traceable execution chain, eight naturally triggered a within-batch low-reward outlier. Table~\ref{tab:difficultprocess} describes only these eight runs: 30 initial candidates included eight low-reward outliers; attribution-guided refill rounds produced 35 scored candidates, and 32 candidates were ultimately retained. Logic-pass rates were 50.00\% for the outliers and 91.43\% for the refill candidates; high-risk rates were 50.00\% and 8.57\%, respectively. The candidate sets are not paired, so their mean-reward difference is not a per-candidate repair gain. This trace does not separately validate the later End-stage batch review.

\begin{table}[t]
\centering
\small
\caption{Variance Mode under a fixed action-history snapshot. Counts describe 30 independent controlled trials, not the anomaly rate of normal operation.}
\label{tab:varianceprocess}
\begin{tabular}{@{}lr@{}}
\toprule
Outcome & Count \\
\midrule
Completed controlled trials & 30 \\
Historical low-reward outliers & 6 \\
Strict repair successes & 3 \\
Valid repairs without reward gain & 1 \\
\bottomrule
\end{tabular}
\end{table}

The 30 trials initialize from the same action-history snapshot but use isolated runtime memory. Six trials produced low historical reward outliers. Three met the strict repair criterion: attribution led to a targeted rollback, the repaired candidate exceeded the detected candidate's reward, and the final candidate passed expression, logic, risk, and label-preservation checks. One further repair was valid but did not improve reward. These controlled counts demonstrate that detection and targeted repair can occur; they do not estimate the anomaly frequency or overall repair rate under ordinary generation.

\section{Limitations and Discussion}\label{sec:limitations}
Although \ours shows relatively stable augmentation gains across multiple logical reasoning and natural-language inference tasks, several aspects of the current method could be further improved. First, \ours involves multi-candidate generation, shared validation, attribution, replenishment, and, when necessary, repair and revalidation. Compared with single-pass generative data augmentation, these steps introduce additional inference and validation overhead. The overall processing cost and latency increase further when more candidates are generated or when an instance triggers replenishment and repair. Future work could reduce this additional cost by avoiding unnecessary candidate generation, refining the conditions that trigger validation, and improving the efficiency of feedback and repair.

Second, the effectiveness of conditional symbolic verification depends on whether natural-language instances can be reliably converted into corresponding logical representations. For instances that can be formalized reliably, Z3 provides an additional symbolic logic check. For instances that cannot, the system skips symbolic verification and retains the semantic-validation path, preventing unreliable formalizations from directly affecting candidate decisions. Nevertheless, the conversion from natural language to logical representations can still be affected by linguistic complexity and semantic ambiguity. Improving the accuracy and applicability of the formalization process therefore remains an important direction for future work.

Finally, historical anomaly control in Variance Mode depends on cross-instance execution records accumulated for the same augmentation action. For newly created actions or actions with limited history, the historical mean and standard deviation may not yet provide a stable statistical reference. When subsequent inputs differ substantially from the existing history, that history may also become less representative of the current execution. The current system mitigates the former risk by deferring anomaly judgments until sufficient history is available. Designing more robust cold-start mechanisms, history-window update strategies, and adaptive statistics for distribution shifts remains an open direction.

\section{Conclusion}\label{sec:conclusion}
We present \ours, a dual-mode quality-control framework for logic-preserving data augmentation. Difficult Mode focuses on the current instance and candidate batch, controlling the quality of the current round through selective retention, batch statistics, candidate-level attribution, replenishment of missing candidates, and final batch-level review at the End node. Variance Mode builds a dynamic reference from the cross-instance history of each augmentation action on top of per-instance diagnosis and execution records, and applies further attribution, targeted feedback, and bounded repair to historically anomalous executions. Both modes share LLM-based semantic validation and enable Z3 symbolic verification when a reliable logical representation is available.

Under the Two-Stage Transfer setting, \ours achieves the highest Accuracy on five of seven tasks and outperforms the no-augmentation BERT baseline on all seven. Controlled ablations show that Memory, Z3, and historical anomaly control play complementary roles within the overall framework. Direct Augmentation, generated-data quality evaluation, and process-level analyses of both modes provide additional evidence for the framework's effectiveness and stability in logic-preserving data augmentation.

Future work will improve the reliability of natural-language formalization, reduce the overhead of multi-candidate generation and multi-round validation, and investigate more robust action-history modeling, cold-start mechanisms, and adaptive update strategies to improve the stability of Variance Mode when history is sparse or the data distribution changes.

\bibliography{references}
\end{document}